\documentclass[sigconf, screen]{acmart}
\AtBeginDocument{%
  }

\setcopyright{acmlicensed}
\copyrightyear{2025}
\acmYear{2025}
\acmDOI{XXXXXXX.XXXXXXX}
\acmConference[MM '25]{ACM Multimedia 2025}{Oct. 27--31,
  2025}{Dublin, Ireland}
\acmBooktitle{MM '25: ACM Multimedia,
 Oct. 27--31, 2025, Dublin, Ireland}
\acmISBN{978-1-4503-XXXX-X/2018/06}

\acmSubmissionID{1787}

\usepackage{algpseudocodex}
\usepackage{algorithm}
\usepackage{multirow}
\begin{document}

\title{Action Unit Enhance Dynamic Facial Expression Recognition}

\author{Feng Liu}
\email{lsttoy@163.com/liu.feng@sjtu.edu.cn}
\orcid{https://orcid.org/0000-0002-5289-5761}
\affiliation{%
  \institution{School of Psychology, Shanghai Jiao Tong University}
  \city{Shanghai}
  \country{China}
}
\authornote {Primary corresponding author}

\author{Lingna Gu}
\email{z5509575@ad.unsw.edu.au}
\orcid{https://orcid.org/}
\affiliation{%
  \institution{University of New South Wales}
  \city{Sydney}
  \country{Australia}
}

\author{Chen Shi}
\email{52205901019@stu.ecnu.edu.cn}
\orcid{https://orcid.org/}
\affiliation{%
  \institution{East China Normal University}
  \city{Shanghai}
  \country{China}
}

\author{Xiaolan Fu}
\email{fuxiaolan@sjtu.edu.cn}
\orcid{https://orcid.org/0000-0002-6944-1037}
\affiliation{%
  \institution{School of Psychology, Shanghai Jiao Tong University}
  \city{Shanghai}
  \country{China}
}
\authornote {Corresponding author}

\renewcommand{\shortauthors}{Feng Liu et al.}

\begin{abstract}
Dynamic Facial Expression Recognition(DFER) is a rapidly evolving field of research that focuses on the recognition of time-series facial expressions. While previous research on DFER has concentrated on feature learning from a deep learning perspective, we put forward an AU-enhanced Dynamic Facial Expression Recognition architecture, namely AU-DFER, that incorporates AU-expression knowledge to enhance the effectiveness of deep learning modeling. In particular, the contribution of the Action Units(AUs) to different expressions is quantified, and a weight matrix is designed to incorporate a priori knowledge. Subsequently, the knowledge is integrated with the learning outcomes of a conventional deep learning network through the introduction of AU loss. The design is incorporated into the existing optimal model for dynamic expression recognition for the purpose of validation. Experiments are conducted on three recent mainstream open-source approaches to DFER on the principal datasets in this field. The results demonstrate that the proposed architecture outperforms the state-of-the-art(SOTA) methods without the need for additional arithmetic and generally produces improved results. Furthermore, we investigate the potential of AU loss function redesign to address data label imbalance issues in established dynamic expression datasets. To the best of our knowledge, this is the first attempt to integrate quantified AU-expression knowledge into various DFER models. We also devise strategies to tackle label imbalance, or minor class problems. Our findings suggest that employing a diverse strategy of loss function design can enhance the effectiveness of DFER. This underscores the criticality of addressing data imbalance challenges in mainstream datasets within this domain. The source code is available at \href{https://github.com/Cross-Innovation-Lab/AU-DFER}{https://github.com/Cross-Innovation-Lab/AU-DFER}.
\end{abstract}




\maketitle

\section{Introduction}
In human communication, facial expression is an important part of non-verbal information, which takes up 60–80\% of communication\cite{non-verbal}. Facial Expression Recognition(FER) is the specialized task in computer vision to classify the emotion in given photos. As more datasets containing video and multiple video frames have been proposed, FER task based on these data can also be called DFER.  

Previous works \cite{AU-vit,10394636,tian-2001} use Action Unit(AU) to improve the performance of static FER models, considering facial expression as a combination of facial action activation. Zhi\cite{zhi-2020} et al. combines AUs-emotion knowledge with 3DCNN, to capture temporal domain feature, yet only 12 AUs were studied and the model only focused on the top 5 AUs. A method to comprehensively and quantitatively analyze the relationship between AU and expression is still lacking.

In the field of psychology, Ekman was the first to develop the use of the Facial Action Coding System(FACS), which is based on facial movement units, for the purpose of distinguishing and recognizing basic human expressions\cite{Ekman1978FacialAC}. This was subsequently employed as a foundation for integrating the OCC theory\cite{OCC}, which led to the establishment of the current domain of expression recognition. Facial Action Units(AU) are used to express the muscle groups that move in the face, representing both the static muscle group configuration and the dynamic movement state of the muscle groups during expression\cite{AU-vit,10394636,tian-2001}. Although numerous studies\cite{10204851,10204765} have been conducted in the field of expression recognition, FER has sought to make predictions about AU. Furthermore, a limited number of studies\cite{Zhi2020ActionUA,9897960,7284869,Jeganathan2022QuantifyingDF} have employed AU with the objective of enhancing the performance of static expression recognition. 

In contrast to the fixed nature of the FER task, the DFER approach permits the analysis of time-dependent data, as it considers the dynamic aspect of expression change. This renders the use of AU a more advantageous approach for the advancement of DFER-related research\cite{10203738,10203585}. In light of the aforementioned considerations, it is evident that the use of Facial Action Units to enhance the performance of basic expression models in the field of dynamic expression recognition warrants a subfield of its own, namely AU-DFER. It is important to acknowledge the limitations of such algorithmic models, which are caused by the qualitative definitions proposed by Ekman\cite{2005Ekman} in the early stages of this field of study. The basic expression categorization, which represents a rough grouping of emotions based on the presence of specific AU, and the inability to analyze the quantitative relationship between AU and expressions in depth in the early days of computer vision technology development, resulted in the inability to utilize the priori knowledge of AU from the psychology of emotion to guide the classification of basic expression emotions.

\begin{figure}[ht!]
  \centering 
  \vspace{-5pt}
  \includegraphics[width=0.9\linewidth]{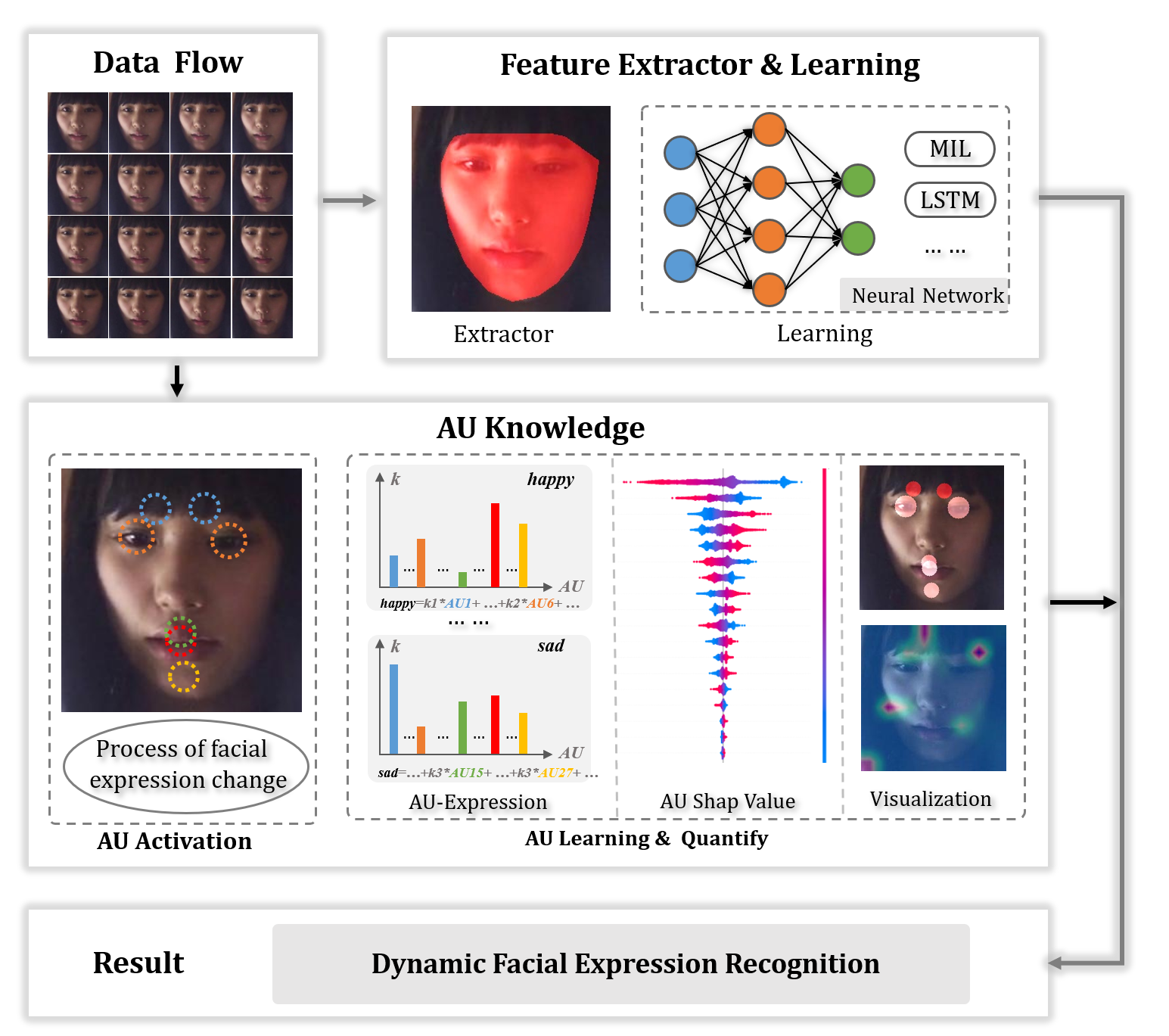}
  \vspace{-5pt}
  \caption{The core architecture design diagram of our AU-enhanced DFER. Highlight the core innovation of this research using AU knowledge to enhance dynamic expression recognition. The figure shows an AU knowledge-enhanced dynamic expression recognition model to enhance recognition using \textit{Sad} expressions as an example. }
  \label{fig:story}

\end{figure}

To address this issue, it is necessary to first analyze the quantitative relationship between AU and expressions and extract it into the a priori knowledge of AU. This should then be used to guide dynamic expression recognition and further injected into mainstream dynamic expression recognition as a plug-and-play middleware. Specifically, 4 existing mainstream DFER datasets(AffWild2, DFEW, FERV39k and RAVDESS) which are employed to analyze the quantitative relationship between AU and expressions. This enables a better design of the loss function for calculating AU loss, which represents the a priori knowledge of AU. Subsequently, the AU loss is integrated with the loss function of the existing open-source top-3 SOTA model. This process culminates in the injection of the a priori knowledge of AU into the dynamic expression recognition base model, thereby enhancing the effectiveness of dynamic expression recognition. Experiments are conducted on DFEW and FERV39k to demonstrate the effectiveness of such a knowledge injection, as shown in Fig. \ref{fig:story}. Furthermore, additional experiments were conducted, employing three knowledge injection strategies: global positive class weight, distinct positive class weight, and minor class positive class weight. These experiments demonstrated that the use of an appropriate strategy can yield optimal results across diverse datasets and base models.

Overall, our contributions can be summarized as follows:
\begin{itemize}
    \item In order to extend the quantitative study of the Action Unit(AU) as defined in the field of emotion psychology, we employ the qualitative study of 4 dominant dynamic expression datasets(AffWild2, DFEW, FERV39k and RAVDESS). The results of this study are then applied to optimize the task in the field of Dynamic Expression Recognition(DFER), with the proposal of a new segmentation, namely AU-DFER.
    \item Our proposed architecture could be implemented on any DFER model without necessitating a change in scale. Furthermore, the method has been tested on 3 dynamic expression datasets, DFEW, FERV39k and MAFW, where it has demonstrated the ability to achieve State-Of-The-Art(SOTA) results while maintaining efficiency.
    \item We investigate the issue of imbalanced samples in the DFER task. The experimental results demonstrate that the AU-DFER is effective architecture in targeting different emotions for effect enhancement.
    \item Our findings corroborate specific quantitative relationship between AU and expression, thereby establishing a robust foundation for further optimisation of Ekman's basic emotion theory in emotional psychology.
\end{itemize}

\section{Related Work}
\subsection{Dynamic Facial Expression Recognition}
DFER is a task to classify the expression of subjects using video data. It is developed from static facial expression recognition, as more video-based datasets have been proposed. The most significant characteristic of this task is that it is based on video sequences rather than single images\cite{why, 10908623}.

In accordance with the essence of video-based data, many DFER approaches focus on temporal and spatial features. Thus, Long Short-Term Memory(LSTM), an architecture that can deal with different temporal relation is widely used for this task\cite{tran2015learning, jiang2020dfew}. Zhao and Liu \cite{Former} proposed a network with a temporal transformer and a spatial transformer, to learn contextual features from both perspectives, resulting in more discriminative facial features being captured. Similarly, Gong et al.\cite{gong-2023} proposed an Enhanced Spatial–Temporal Learning Network(ESTLNet) to gain more robustness. Gowda et al.\cite{10581905} treat image data and video data as 2 modalities, and proposed Facial-Emotion Adapter(FE-Adapter) to achieve efficient fine-tuning. Chen et al.\cite{chen-2024} proposed a method to learn geometry structure knowledge under Euclidean and Hyperbolic spaces, named MGDL, and applied it in both temporal and spacial dimensions. Also, Li et al.\cite{li-2023} introduced an intensity-aware loss in the process of training their model.
\begin{figure*}[ht]
  \centering
  \includegraphics[width=0.85\linewidth]{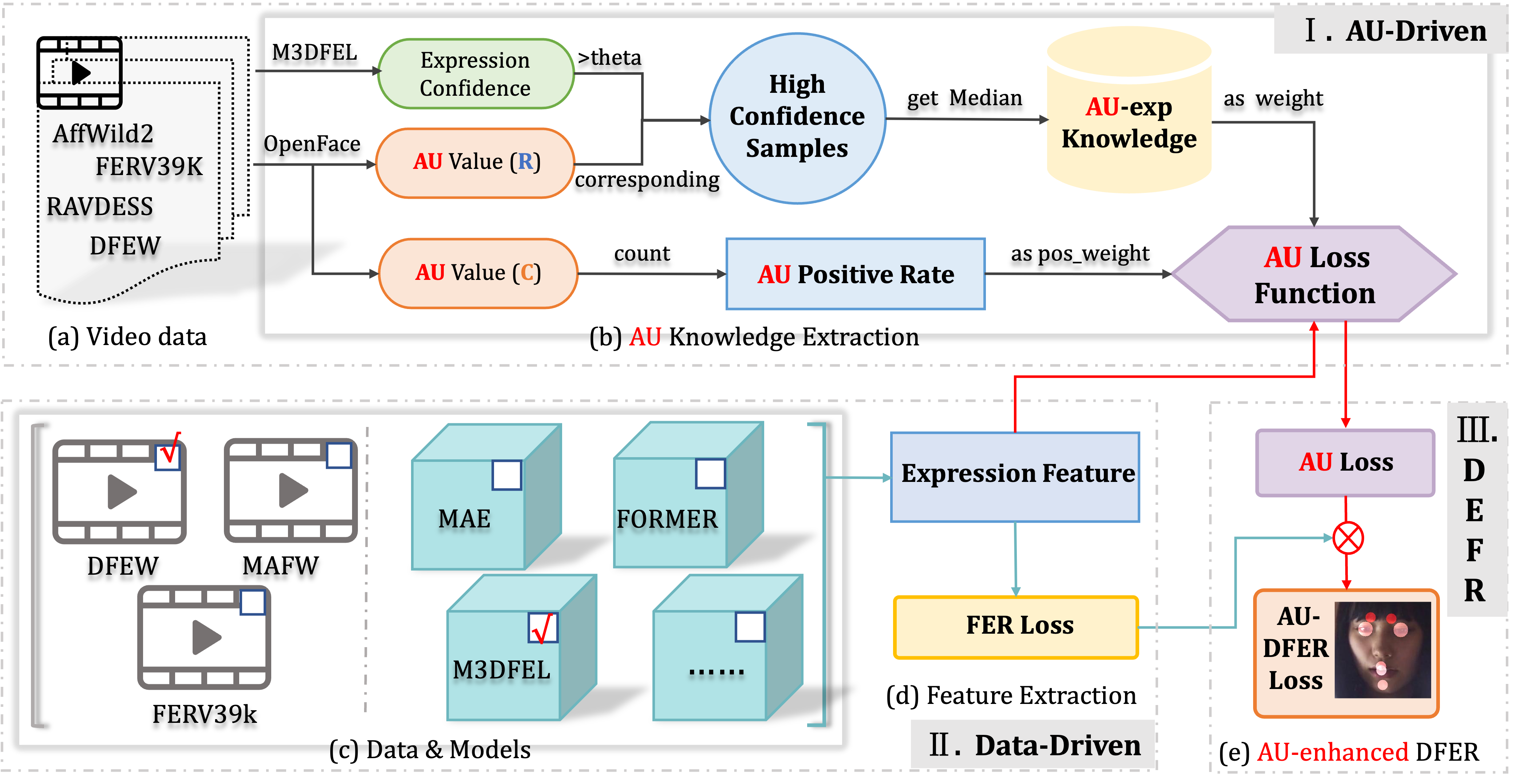}
  \vspace{-5pt}
  \caption{An overview of the proposed AU-DFER architecture.(\uppercase\expandafter{\romannumeral1}) AU-Driven loss design (\uppercase\expandafter{\romannumeral2}) Data-Driven Model Selection (\uppercase\expandafter{\romannumeral3}) combination of Knowledge and existing model. }
  \label{fig:architecture}
\end{figure*}

Despite the rapid development of this field, there are still some problems unsolved, such as imbalanced data categories in terms of datasets, and lack of robustness for models. Some recent works made an effort to tackle these problems with some methods that highly rely on data. Sun et al. conveyed that massive unlabeled data should be used, and proposed self-supervised models such as MAE-DFER\cite{MAE-DFER} and SVFAP\cite{sun-2024}. Chen et al.\cite{chen-2024} use information from other modality, such as text, to enhance the performance. 
Wang et al.\cite{why} put forward that DFER task should be considered as a weak supervised learning problem due to the existence of inexact labeled frames, which reveals the essence in temporal relation. Similarly, the majority of extant studies have been conducted from the vantage point of samples and labels. In contrast, we put forth a novel perspective based on emotional psychology to enhance DFER through the utilization of a priori knowledge regarding Dynamic AUs associated with facial expressions. We thus propose a novel insight to enhance the efficiency of spatial feature extraction with AUs, adding quantified AU-expression knowledge to existing approaches.

\subsection{Facial Action Coding System}
Facial Action Coding System(FACS) is a technique of part-based methods where facial parts are considered to analyze human emotion\cite{Sarangi_Panda_Mishra_Mishra_Majhi_2022}. It describes facial activities in terms of anatomically based Action Units(AU)\cite{8802914}. Studies \cite{parr-2010} have conveyed that such a system is standardized, thus can be applied on people from different ethnic groups and cultural backgrounds. Jeganathan et al.\cite{jeganathan-2022} discovered the difference in AU activation in response to emotion simulations between depressed individuals and healthy people.

Since it was proposed by Ekman and Friesen\cite{ekman_2005_what}, FACS has been frequently used as a tool to support FER. Chen and Wang used facial landmark to help locate AUs, and generated Gaussian heatmaps and crop AU regions of interest(RoI) to support FER\cite{10394636}. These works are based on image data, ignoring the fact that both AU and emotion are a dynamic process, which indicates that AU can actually better support DFER than to support FER.

What's more, the relationship between certain Facial Action Unit and emotion remains qualitative according to the basic emotion theory of Ekman, making it impossible to discover its full potential of assisting DFER. Tian et al.\cite{tian-2001} focused on the geologic distribution of AUs on human face and different combinations of AUs corresponding to different expressions, and proposed several combinations of AUs to suggest emotion. Shang et al.\cite{shang-2024} developed a multitask model to achieve AU detection and FER. Even though some works use Facial Action Unit as auxiliary to benefit facial expression recognition, the method adopted is simply adding facial action unit label, or geometric knowledge of AU distribution to the training process, ignoring the fact that different AUs contribute differently to various emotions. Also, human annotation inevitably introduces subjectiveness to datasets. 

To tackle the problem mentioned above, we conduct our study on 4 different datasets with AU annotation, to discover the relationship between AU and DFER in a quantitative by objective way.

\section{Methodology}

\subsection{AU-expression Knowledge Extraction}
To better guide DFER task with AU-expression knowledge, it is necessary to analyze quantitatively. In order to gain the quantitative relationship between Facial AUs and expression, we conduct the following experiment.

As shown in Fig.\ref{fig:architecture}(a), we obtain the AU prediction for every single frame of 4 DFER datasets with OpenFace Ver.2.2.0.
For a video sample $S$ with n frames, extract 17 facial AUs  with OpenFace, as a regression task.
For each frame, there is a corresponding vector, marked as $\mathrm{AUr_{Si}(1\leq i <n)}$, $\mathrm{AUr_{Si}} \in \mathbb{R}^{17}$,
\vspace{-1pt}
\begin{equation}
    {{AUr}}_{ij} = O_r(\text{{S}}_i), \quad i = 1, 2, \ldots, n \quad j = 1, 2, \ldots, 17,
\end{equation}
where $O_r$ denotes using OpenFace to extract AUs as a regression task and sample-wise linear interpolation for 0 values, while ${{AUr}}_{ij} \in [0,5]$ is the result. We also get the values of AU as pseudo labels for training datasets, represented as 
\vspace{-1pt}
\begin{equation}
    {{AUc}}_{ij} = O_c(\text{{S}}_i), \quad i = 1, 2, \ldots, n \quad j = 1, 2, \ldots, 18,
\end{equation}
where $O_c$ denotes using OpenFace to extract AUs as a classification task, while ${{AUc}}_{ij}\in \{0,1\}$ is the result.

We illustrate this using M3DFEL\cite{why} as an example of an AU-enhanced DFER approach. This approach gain the prediction of expression for each frame, marked as $\mathrm{E_{Si}(1\leq i <n)}$, $\mathrm{E_{Si}}\in \mathbb{R}^{7}$,
\begin{equation}
E_i = \text{{M3DFEL}}(\text{{S}}_i), \quad i = 1, 2, \ldots, n.
\end{equation}

Due to the existence of inexact label in DFER, we use original M3DFEL to generate the expression pseudo label and set a threshold of the predicted value of target expression. For each frame, labels can be represented with an integer, written as $\mathrm{l_{Si}(l\in \{0,1,2,3,4,5,6\})}$. We compare the corresponding prediction of target expression with a threshold $\mathrm{\theta}$, only collecting the frames that satisfies$\mathrm{E_{Si}[j]>\theta, j = l_{Si}}$. By doing so, we eliminate the potential bias and pick out the reliable samples to gain AU-expressions knowledge.

Now we have a set of frames that can be grouped with expression labels, and we tally up the median, gaining a knowledge matrix $\mathrm{K}$ on each dataset, $\mathrm{K}\in \mathbb{R}^{18\times7}$, containing the relationship between every single AU and expression. We first minus each number in this matrix with the average of maximum and minimum, so that the range of values is symmetric about 0. Then we normalize it using sigmoid. It is worth reminding that OpenFace does not provide a regression result of AU28, while it contains AU28 as the result of a classification task, a 0-1 value which we later use as pseudo label, so we use the average of other other 17 AUs instead of calculating median.

\begin{figure}[ht]
  \vspace{-5pt}
  \centering
  \includegraphics[width=.9\linewidth]{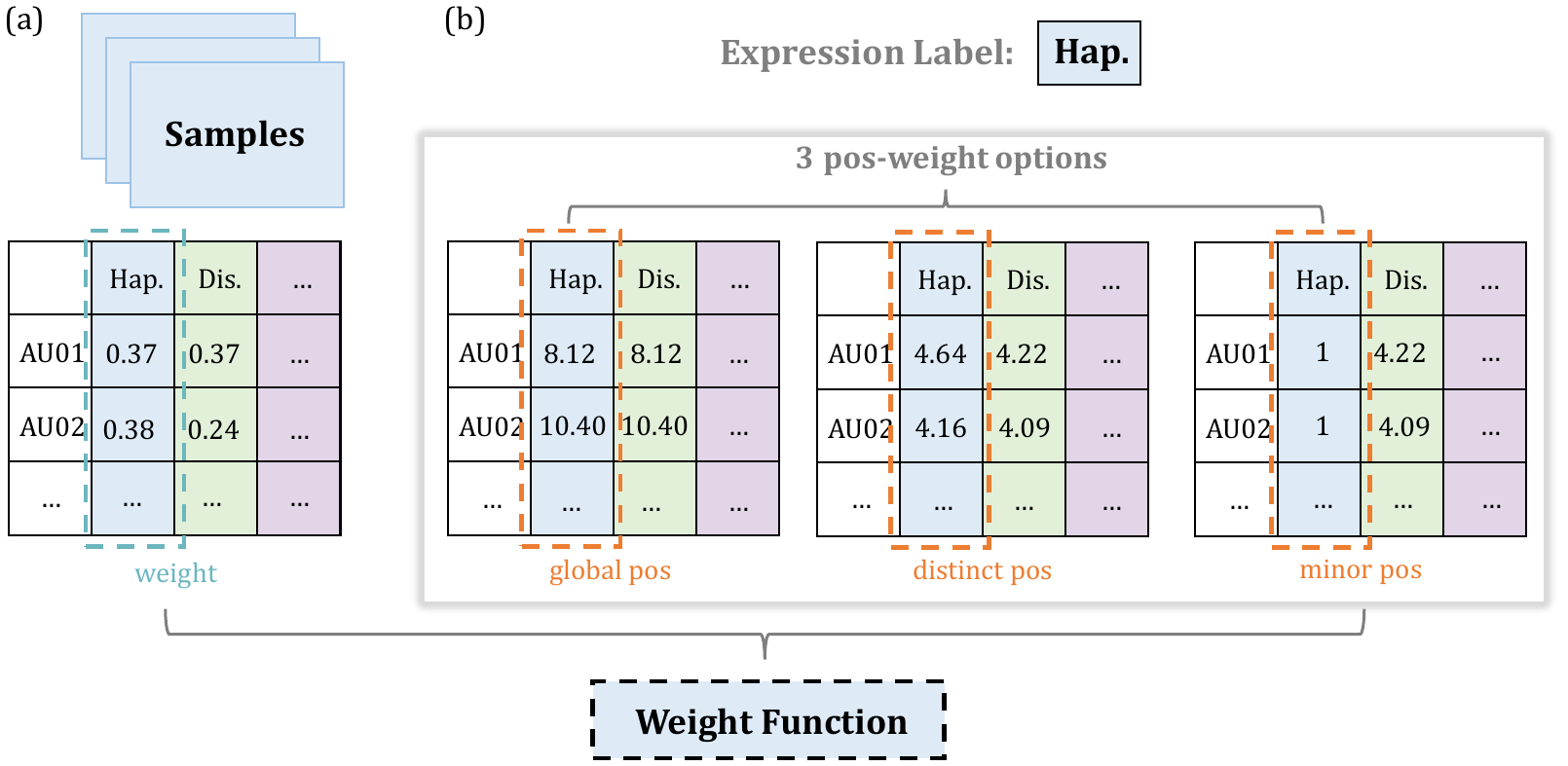}
  \vspace{-5pt}
  \caption{Design conceptualization of the AU loss function.}
  \label{fig:loss}
\end{figure}

After adding up the result on 4 datasets,${S = \sum_{i=1}^{4} K_i}$, $\mathrm{S}\in \mathbb{R}^{18\times7}$, we find that the sum is till in the range of [0,5], exactly between the value of AU given by OpenFace, in order to gain a more discriminating distribution, we then minus all the sums with 2.5 and then use sigmoid to normalize it. When applying it to the loss function, we multiply it with 5 to keep the value in the same range as OpenFace provide. The matrix is used as weight in loss function, as shown in Fig. \ref{fig:loss}.

\subsection{Dynamic Facial Feature from DFER Approaches}
\label{sec:insta_gener}

Our AU-expression knowledge can be implemented to any DFER network. The M3DFEL \cite{why} from CVPR 2023 exemplifies this approach, which is employed in our experiments and depicted in Fig. \ref{fig:architecture}(c).  After feature extraction, or before classification, we add an AU classifier to detect AU, in addition to expression. The feature extractor takes a group of video frames as input, and outputs a vector with 1024 elements, which is the input of expression classifier. The expression classifier returns a 7-element vector, given the number of expression class is 7. We simply add an AU classifier, taking the 1024-element feature vector as input as well, and returning a result of AU detection. Typically, the return value is an 18-element vector, as 18 AUs are labeled. For each video sample, we add up the 0-1 values from OpenFace, and compare the sum with the number of frames, then we have the AU label for the whole video. For a video with n frames,
\begin{equation}
    Y_{AU_i} = 
    \begin{cases}
        1 (\sum_{j=1}^{j=n}{AUc_{ij}}\geq0.5n)\\
        0 (\sum_{j=1}^{j=n}{AUc_{ij}}<0.5n)\\
    \end{cases}
\end{equation}

\subsection{AU Feature Injection with Loss Combination}
\label{sec:insta_gener}
To utilize the AU-expression knowledge, we modify the loss function to let the result of AU prediction better guide the feature extractor.

The expression loss can be calculated as following:
\begin{equation}
    L_e = -\frac{1}{N} \textstyle\sum_{i=1}^{N} \sum_{j=1}^{7} y_{ij} \log(5\cdot p_{ij})
,
\end{equation}
where\( y_{ij} \) is the annotation for the $j$-th expression of $i$-th sample, and \( p_{ij} \) is the prediction of the $j$-th expression of $i$-th sample.

Due to the strong correlation between AUs and facial expressions at the representational level, the attainment of favorable AU classification outcomes is indicative of the model's enhanced capability to discern more resilient facial characteristics. Conversely, the realization of favorable emotion classification outcomes serves to augment the model's aptitude for precise AU recognition, thereby conferring a state of complementary advantage. In order to ascertain the optimal methodology for knowledge injection, three different approaches to calculating AU loss are employed, as Fig.\ref{fig:loss}(b2) demonstrates. In essence, the utilized method is BCEWithLogitsLoss, with the differentiating factor being the attribute utilized. In light of the imbalance across categories, we define happy, sad, angry, and neutral as major classes, each comprising more than one-seventh of all samples. The remaining three are referred to as minor classes. The attribute weight is the knowledge gained in section 3.2, while the positive class weight varies.

In the first method, the positive class weight for 7 classes are the same, for the $j$-th AU, we have
\begin{equation}
    pw_j = \frac{c-\sum_{k=1}^{k=c}{Y_{AU_{kj}}}}{\sum_{k=1}^{k=c}{Y_{AU_{kj}}}},
    \label{eq:global}
\end{equation}
here c is the number of all samples in a dataset.
In the second method, we calculate positive class weight respectively for each expression class, for the $j$-th AU and $i$-th expression
\begin{equation}
    pw_{ij} = {\frac{c_i-\sum_{k=1}^{k=c_i}{Y_{AU_{kj}}}}{\sum_{k=1}^{k=c_i}{Y_{AU_{kj}}}}},
    \label{eq:distinct}
\end{equation}
where $c_i$ is the number of all samples labeled as class $i$.

In the third method, we only apply positive class weight on minor classes,
\begin{equation}
    pw_{ij} = 
    \begin{cases}
        1 \text{ (i in major class)}\\
        {\frac{c_i-\sum_{k=1}^{k=c_i}{Y_{AU_{kj}}}}{\sum_{k=1}^{k=c_i}{Y_{AU_{kj}}}}}\text{ (i in minor class)}\\
    \end{cases}.
    \label{eq:minor}
\end{equation}

Based on the weight and positive class weight attributes above, we have

\begin{equation}
  \resizebox{0.45\textwidth}{!}{$   L_{AU} = -\frac{1}{N} \sum_{i=1}^{N} \left( k_{j} \cdot \left( pw \cdot Y_{AU_i} \cdot \log(\sigma(au_i)) + (1 - Y_{AU_i}) \cdot \log(1 - \sigma(au_i)) \right) \right) $},
\end{equation}
where $\sigma$ denotes sigmoid function, $N$ is the total sample number, $k_j$ is the $j$-th row of knowledge matrix $K$(j is the expression label), and $au_i$ is the prediction of AU for sample $i$.

The combination of 2 loss functions are depicited in Fig. \ref{fig:architecture}(e). Specifically, we assign different weights to the two types of losses,

\begin{equation}
    L = (1-\lambda) * L_e + \lambda * L_{AU}\;\;.
\end{equation}

We conduct a simple experiment to study about the best possible ratio and will discuss about it later.

\begin{figure}
\centering
\includegraphics[width=\linewidth]{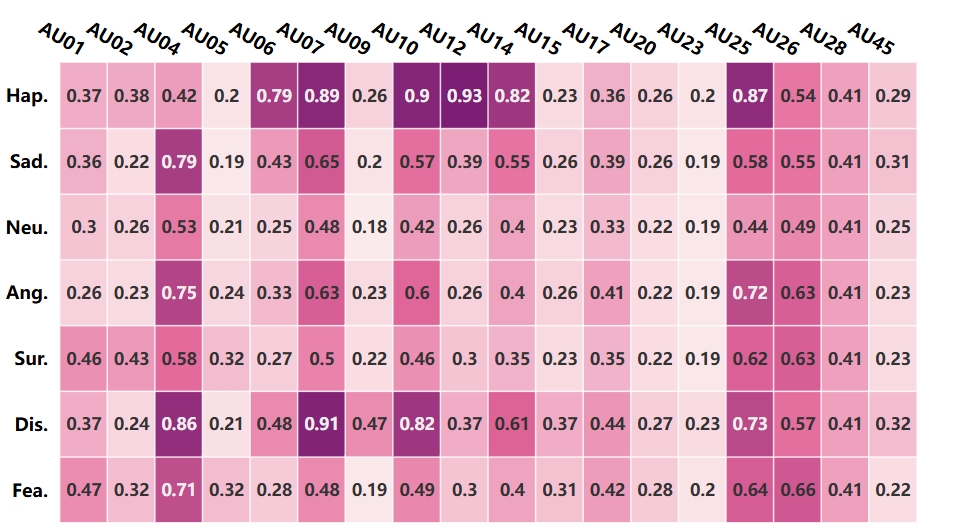}
\caption{Visualization of the AU expression knowledge acquired from 4 mainstream DFER datasets.}
\label{fig:knowledge}
\end{figure}

\begin{figure*}
  \centering
  \includegraphics[width=\linewidth]{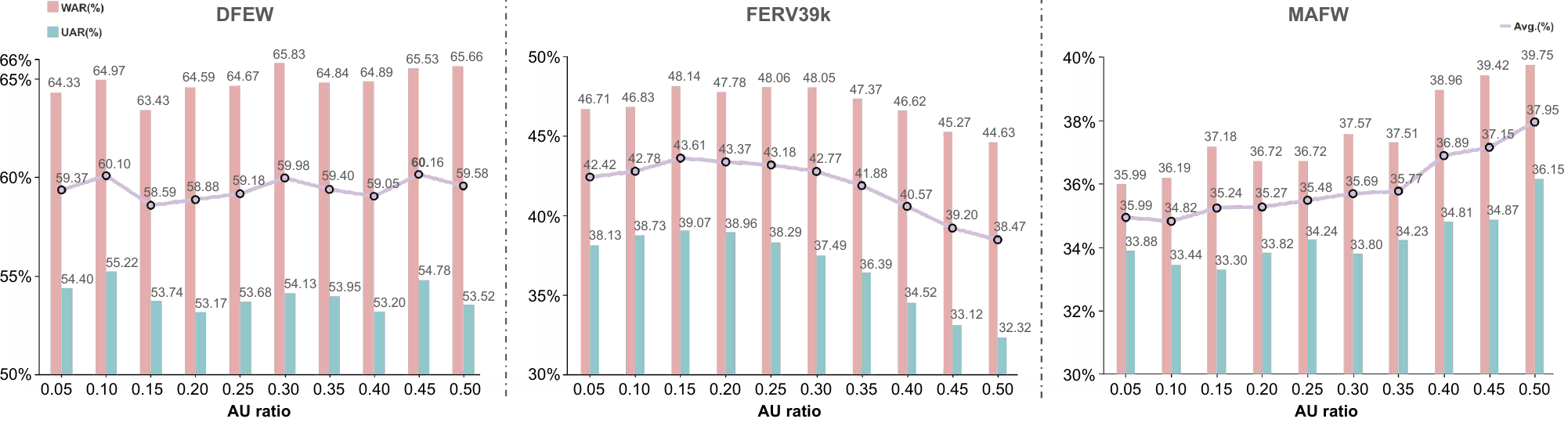}
  \caption{
  Ablation Study of the AU-expression loss ratio using M3DFEL with AU-DFER on 3 datasets.}
  \label{fig:ablat}
\end{figure*}

\section{Experiments}

\subsection{Datasets}

We conduct our AU-expression knowledge rediscovery experiment on 4 publicly available video-based datasets: AffWild2\cite{AffWild2}, DFEW\cite{jiang2020dfew}, FERV39k\cite{wang2022FERV39k}, and RAVDESS\cite{Livingstone_Russo_2018}. AffWild2 contains 2613000 video frames downloaded from Youtube with expression annotated. DFEW consists of more than 16,000 video clips from movies, and some extreme interference exists in its samples. FERV39k includes 38,935 video clips which have been first divided into 4 scenarios and then subdivided into 22 scenes labeled with 7 basic expressions. RAVDESS is a collection of videos in which 24 professional actors and actresses say two lexically-matched statements.

We train and test Top-3 models on 3 datasets: DFEW, FERV39k and MAFW\cite{MAFW}. MAFW contains 10045 video clips with expression annotation from diverse screenmedia.

\subsection{AU Feature as Injection Knowledge}

As mentioned in section 3.2, we extract AU-expression knowledge to assist DFER task. Fig. \ref{fig:knowledge} shows the values we use as weight in our architecture. As OpenFace only provides classification results for AU28, we do not have access to the regression results. Consequently, we use the average of all other AUs instead. The heatmap reveals the intensity of each AU in different expressions from 4 existing mainstream dynamic expression recognition datasets(AffWild2, DFEW, FERV39k and RAVDESS). Some AUs, such as AU05, AU23, and AU26, tend to be equally active in different expressions, while others, like AU07, AU12 and AU14, are extremely active in one or two expressions. This might result from various features of facial muscles and nerves.

\subsection{Implementation Details}
All of our experiments are conducted using GeForce RTX 4090D GPU. As to software, all models use PyTorch-GPU. For fair comparison, we run all baseline models on the same device as our own method.

For M3DFEL\cite{why}, the vanilla R3D18 is employed as backbone, learning rate is initialized to $5\times10^{-4}$, with a minimum of $5\times10^{-6}$, and the weight decay is set to 0.05. The overall epoch number is 100. As to MAE-DFER\cite{MAE-DFER}, a pre-trained model on VoxCeleb2 \cite{VoxCeleb2} is adopted. During the finetuning process, an AdamW with $\beta_1$ = 0.9 and $\beta_2$ = 0.999 is used as the optimizer, and the learning rate is initialized to $10^{-3}$. Similarly, for Former-DFER\cite{Former}, weight decay is $10^{-4}$, learning rate is 0.01, and total epoch number is 100. All aforementioned configurations of approaches are identical to the official code release.

\subsection{Main Baselines}
Following are some of the baselines we compare with in our experiment. We modify the first 3 of them, to show the effectiveness of AU-expression knowledge and to prove the flexibility as a plug-in module.

Former-DFER(MM 2021)\cite{Former} spatial transformer and temporal transformer learn features from spacial and temporal perspectives respectively.

MAE-DFER(MM 2023)\cite{MAE-DFER} developed from Vision Transformer(ViT), local-global interaction Transformer encodes features efficiently.

M3DFEL(CVPR 2023)\cite{why} proposed a Multi-3D Dynamic Facial Expression Learning framework, which utilizes Multi-Instance Learning to handle inexact labels. 

S2D(TAC 2024)\cite{S2D} add Temporal-Modeling Adapters and landmark-aware feature extractor to a ViT and Multi-View Complementary Prompters(MCPs)-based Static Expression Recognition model.

SVFAP(TAC 2024)\cite{SVFAP} a self-supervised model with a temporal pyramid and a spatial bottleneck Transformer to handle spatio-temporal redundancy.

\section{Results}

\subsection{Comparison with the SOTA Methods}

Tab. \ref{tab:comp_2} shows WAR and UAR before and after adding AU-expression knowledge to 3 models on DFEW, FERV39k and MAFW. Our proposed architecture improves all models on all datasets. Both WAR and UAR increases by around 1 percent. Especially, our proposed method works on MAFW, which suggests the AU-expression knowledge is universal across datasets. To further explore the contribution of AU knowledge, we add it to existing models and compare the performance. Experiments show that AU knowledge can uniformly improve the accuracy of DFER models. The result is shown in Tab. \ref{tab:comp_DFEW}. The result shows our proposed architecture boosts all 3 models, and FLOPs remains the same. The result shows AU-expression knowledge can enhance DFER model without cost increase. Taking a closer look at accuracy of 7 different categories, our method is more efficient on optimizing the performance on minor classes. Typically, M3DFEL's accuracy of 'Disgust' increased from 0.00\% to 10.34\%. This suggests that such a method has the potential to deal with imbalanced data.

\begin{table}[t]
    \centering
    \footnotesize
    \setlength{\tabcolsep}{4pt}
    \renewcommand{\arraystretch}{1.1}
    \begin{tabular}{l|cc|cc|cc}
        \toprule
        \textbf{Method} & \multicolumn{2}{c}{\textbf{DFEW (\% ↑)}} & \multicolumn{2}{c}{\textbf{FERV39k (\% ↑)}} & \multicolumn{2}{c}{\textbf{MAFW (\% ↑)}} \\
        & \textbf{WAR} & \textbf{UAR}  & \textbf{WAR} & \textbf{UAR} & \textbf{WAR} & \textbf{UAR} \\
        \midrule
        C3D \cite{tran2015learning} & 53.54 & 42.74  & 31.69 & 22.68 &31.17 &42.25\\
        P3D \cite{qiu2017learning} & 54.47 & 43.97  & 33.39 & 23.20 &- &-\\
        I3D \cite{carreira2017quo} & 54.27 & 43.40 & 38.78 & 30.17 &- &-\\
        R(2+1)D18 \cite{tran2018closer}& 53.22 & 42.79 & 41.28 & 31.55 &-&-\\
        3D ResNet18 \cite{hara2018can} & 54.98 & 44.73 & 37.57 & 26.67 &-&-\\
        ResNet18+LSTM \cite{jiang2020dfew} & 53.08 & 42.86 & 63.85 & 51.32 &28.08 &39.38\\
        EC-STFL \cite{jiang2020dfew} & 56.51 & 45.35 & - & - &- &- \\
        STT \cite{ma2022spatio} & 66.45 & 54.58 & 48.11 & 37.76 &- &- \\
        DPCNet \cite{wang2022dpcnet} & 66.32 & 55.02 & - & -  &- &-\\
        NR-DFERNet \cite{li2022nr} & 68.19 & 54.21 & 45.97 & 33.99 &- &- \\
        SVFAP-S \cite{SVFAP} & 72.67 & 60.45 & 51.34 & 41.19 &39.82	&53.89 \\
        SVFAP-B \cite{SVFAP} & 74.27 & 62.83 & 52.29 & 42.14 & 41.19 & 54.28\\
        S2D \cite{S2D} & 74.81 & 65.45 & 46.21 & 43.97 & 43.40 & 57.37 \\
        FRU-Adapter \cite{FRU-Adapter} & 66.02 & 76.96 & 38.65 & 50.12 & 42.80 & 57.83 \\
        RDFER \cite{rdfer} & 69.72 & 56.93 & 48.60 & 36.47 & - & - \\
        \midrule
        †M3DFEL* \cite{why} & 64.33 & 52.67 & 46.05 & 38.53 &36.39 &34.01 \\
        \;\;\textbf{(+)AU-DFER} & \textbf{65.53} & \textbf{54.78} & \textbf{48.15} & \textbf{38.75} &\textbf{39.75} &\textbf{36.15}\\
        †MAE-DFER* \cite{MAE-DFER} & 73.73 & 61.07 & 52.06 & 42.19 & 48.71 &46.31\\
        \;\;\textbf{(+)AU-DFER} & \textbf{75.01} & \textbf{61.87}  & \textbf{52.45} & \textbf{42.48} &\textbf{49.44} &\textbf{47.41} \\
        †Former-DFER* \cite{Former} & 64.16 & 52.49 & 47.09 & 37.24 & 28.89 & 19.44\\
        \;\;\textbf{(+)AU-DFER} & \textbf{65.16} & \textbf{54.02} & \textbf{47.75} & \textbf{38.18} &\textbf{29.13} & \textbf{20.98}\\
        \bottomrule
    \end{tabular}
    \caption{Comparison of various DFER approaches on 3 mainstream datasets with and without AU-DFER. †Results based on local replication. *MAFW result based on 7-class.}
    \label{tab:comp_2}

\end{table}

\begin{table*}[!ht]

\centering
\footnotesize 
\setlength{\tabcolsep}{4pt} 
\renewcommand{\arraystretch}{1.1} 
\begin{tabular}{l|ccccccc|cc|cc}
\toprule
\multirow{2}{*}{\textbf{Method}} & \multicolumn{7}{c}{\textbf{Accuracy of Each Expression (\% ↑)}} & \multicolumn{2}{c}{\textbf{Metrics (\% ↑)}} & \multirow{2}{*}{\textbf{Flops (G ↓)}} \\
\cmidrule(lr){2-8} \cmidrule(lr){9-10} 
& \textbf{Hap.} & \textbf{Sad.} & \textbf{Neu.} & \textbf{Ang.} & \textbf{Sur.} & \textbf{Dis.} & \textbf{Fea.} & \textbf{WAR} & \textbf{UAR}  & \\
\midrule
C3D \cite{tran2015learning} & 75.17 & 39.49 & 55.11 & 62.49 & 45.00 & 1.38 & 0.51 & 53.54 & 42.74 & 38.57 \\
P3D \cite{qiu2017learning} & 74.85 & 43.40 & 54.18 & 60.42 & 40.99 & 0.69 & 23.28 & 54.47 & 43.97 & - \\
I3D \cite{carreira2017quo} & 78.61 & 44.19 & 56.69 & 55.87 & 45.88 & 2.07 & 20.51 & 54.27 & 43.40  & 6.99 \\
R(2+1)D \cite{tran2018closer} & 79.67 & 39.07 & 57.66 & 50.39 & 48.26 & 3.45 & 21.06 & 53.22 & 42.79 & 42.36 \\
3D ResNet18 \cite{hara2018can} & 73.13 & 48.26 & 50.51 & 64.75 & 50.10 & 0.00 & 26.39 & 54.98 & 44.73 & 8.32 \\
ResNet18+LSTM \cite{jiang2020dfew} & 78.00 & 40.65 & 53.77 & 56.83 & 45.00 & 4.14 & 21.62 & 53.08 & 42.86 & 7.78 \\
EC-STFL \cite{jiang2020dfew} & 79.18 & 57.85 & 60.98 & 46.15 & 2.76 & 21.51 & 56.51 & 45.35 & 50.93 & 8.32 \\
STT \cite{ma2022spatio} & 87.36 & 67.90 & 64.97 & 71.24 & 53.10 & 3.49 & 34.04 & 66.45 & 54.58 & - \\
DPCNet \cite{wang2022dpcnet} & - & - & - & - & - & - & - & 66.32 & 55.02 & 9.52 \\
NR-DFERNet \cite{li2022nr} & 88.47 & 64.84 & 70.03 & 75.09 & 61.60 & 0.00 & 19.43 & 68.19 & 54.21 & 6.33 \\
SVFAP-S \cite{SVFAP} & 92.39 & 74.92 & 70.40 & 76.90 & 62.70 & 8.28 & 37.58 & 72.67 & 60.45 & 18.00 \\
SVFAP-B \cite{SVFAP} & 93.13 & 76.98 & 72.31 & 77.54 & 65.42 & 15.17 & 39.25 & 74.27 & 62.83 & 44.00 \\
S2D \cite{S2D} & 93.95 & 78.35 & 70.25 & 78.00 & 61.88 & 25.52 & 50.22 & 74.81 & 65.45 & - \\
FRU-Adapter \cite{FRU-Adapter} & 66.02 & 76.96 & 38.65 & 50.12 & 61.88 & 25.52 & 50.22 & 74.81 & 65.45 & - \\
RDFER \cite{rdfer} & 89.69 & 69.22 & 70.18 & 71.47 & 62.08 & 0.69 & 28.71 & 69.73 & 56.93 & - \\
\midrule
†M3DFEL \cite{why} & 85.28 & 70.71 & 58.05 & 62.53 & 63.95 & 0.00 & 28.18 & 64.33 & 52.67 & 1.66 \\
\;\textbf{(+)AU-DFER} & \textbf{92.02} & 62.80 & \textbf{60.11} & \textbf{66.90} & 58.16 & \textbf{10.34} & \textbf{33.15} & \textbf{65.53} & \textbf{54.78} & \textbf{1.66}  \\
†MAE-DFER \cite{MAE-DFER} & 95.50 & 73.88 & 73.97 & 75.86 & 65.31 & 10.34 & 32.60 & 73.73 & 61.07 & 26.11 \\
\;\textbf{(+)AU-DFER} & 94.89 & \textbf{75.99} & \textbf{75.09} & \textbf{77.93} & \textbf{68.03} & 6.90 & \textbf{34.35} & \textbf{75.01} & \textbf{61.87} & 26.11 \\
†Former-DFER \cite{Former} & 83.80 & 60.88 & 66.47 & 65.58 & 55.96 & 4.14 & 30.60 & 64.16 & 52.49 &8.32 \\
\;\textbf{(+)AU-DFER} & 83.80 & \textbf{65.98} & 64.13 & 65.36 & \textbf{59.06} & \textbf{4.83} & \textbf{35.00} & \textbf{65.16} & \textbf{54.02} & 8.32 \\
\bottomrule
\end{tabular}
\caption{Comparison with methods on DFEW, including top-3 DFER approaches with AU-DFER. †Results based on local replication.}
\label{tab:comp_DFEW}
\vspace{-15pt}
\end{table*}

\begin{table*}[t]
\centering
\footnotesize 
\setlength{\tabcolsep}{4pt} 
\renewcommand{\arraystretch}{1.1} 
\begin{tabular}{l|c|ccccccc|cc}
\toprule
Method \textbf{(+)AU-DFER} & \textbf{AU Strategy(\% ↑)} & \textbf{Hap.} & \textbf{Sad.} & \textbf{Neu.} & \textbf{Ang.} & \textbf{Sur.} & \textbf{Dis.} & \textbf{Fea.} & \textbf{WAR} & \textbf{UAR} \\
\midrule
\multirow{4}{*}{M3DFEL \cite{why}} 
    & Baseline & 85.28 & \textbf{70.71} & 58.05 & 62.53 & \textbf{63.95} & 0.00 & 28.18 & 64.33 & 52.67 \\
    & \textbf{global pos.} & \textbf{92.02} & 62.80 & \textbf{60.11} & 66.90 & 58.16 & 10.34 & 33.15 & \textbf{65.53} & \textbf{54.78}  \\
    & distinct pos. & 86.50 & 65.44 & 55.99 & \textbf{68.28} & \textbf{63.95} & 0.00 & \textbf{34.25} & 64.80 & 53.49  \\
    & minor pos. & 88.75 & 65.44 & 61.05 & 64.83 & 55.44 & \textbf{17.24} & 24.31 & 64.16 & 53.87 \\
\midrule
\multirow{4}{*}{MAE-DFER \cite{MAE-DFER}} 
    & Baseline & \textbf{95.50} & 73.88 & 73.97 & 75.86 & 65.31 & \textbf{10.34} & 32.60 & 73.73 & 61.07 \\
    & global pos. & 95.09 & 74.14 & \textbf{75.66} & 76.32 & 67.35 & 3.45 & 33.15 & 74.37 & 60.74 \\
    & \textbf{distinct pos.} & 94.89 & \textbf{75.99} & 75.09 & \textbf{77.93} & \textbf{68.03} & 6.90 & 34.25 & \textbf{75.01} & \textbf{61.87} \\
    & minor pos. & 94.48 & 74.41 & 74.72 & 77.24 & 64.63 & \textbf{10.34} & \textbf{35.91} & 74.20 & 61.68 \\
\midrule
\multirow{4}{*}{Former-DFER \cite{Former}} 
    & Baseline & 83.80 & 60.88 & \textbf{66.47} & 65.58 & 55.96 & 4.14 & 30.60 & 64.16 & 52.49\\
    & global pos. & \textbf{85.15} & 60.70 & 64.43 & \textbf{66.48} & 57.49 & \textbf{6.21} & 33.15 & 64.53 & 53.37 \\
    & \textbf{distinct pos.} & 83.80 & \textbf{65.98} & 64.13 & 65.36 & \textbf{59.06} & 4.83 & \textbf{35.00} & \textbf{65.16} & \textbf{54.02} \\
    & minor pos. & 84.70 & 63.46 & 64.14 & 65.39 & 57.79 & 4.83 & 33.48 & 64.67 & 53.40 \\
\bottomrule
\end{tabular}
\caption{Comparison of various weighting strategies on the DFEW dataset for top-3 DFER approaches (+) AU-DFER. Both WAR and UAR are used as primary metrics, with the Avg. metric representing their mean values to give a balanced perspective on the impact of different AU weighting parameters across datasets and approaches. AU Strategy 'global pos.' means using Eq. \ref{eq:global} as positive class weight, while 'distinct pos.' stands for using Eq. \ref{eq:distinct} as positive class weight, and 'minor pos.' refers to adopting Eq. \ref{eq:minor}.}
\label{tab:methDFEW}
\vspace{-15pt}
\end{table*}

\subsection{AUs' Various Weighting Strategies}

\textbf{Evaluation of AU enhancing approaches.} We conduct an experiment to find a best way of combining AU knowledge with 3 existing approaches on DFEW. All the three approaches are introduced in section 3.4, and the results on DFEW are presented in Tab. \ref{tab:methDFEW}.

From the Tab. \ref{tab:methDFEW}, it is clear that AU-enhanced approaches outperform all approaches without AU on DFER, and both WAR and UAR are improved. In addition, using distinct positive class weight and minor positive class weight tend to lead to a best result on 3 minor classes, which indicates that adjusting targeted pos\_weight can benefit the performance on minor classes. And this table also shows that for M3DFEL, the best method is using global positive class weight, while applying distinct positive class weight works best for MAE-DFER and Former-DFER. 
The same experiment is conducted on FERV39k, and we also explore the same model on different datasets. The results indicate that both DFER models and datasets could impact the best strategy of AU-expression knowledge injection, and approaches might be an important factor.

\textbf{Evaluation of different AU-expression loss ratio.} We conduct this experiment to find out the best ratio of AU loss and expression loss when adding up together, using the same positive loss method.
The result is presented in Fig. \ref{fig:ablat}, the experiment shows that the best ratio on different datasets varies. As AU loss :expression loss ratio increases, both WAR and UAR fluctuate on DFEW, and the performance first increase, then decrease on FERV39k. Both WAR and UAR continue to improve on MAFW as the ratio increases. 

\subsection{Visualization}
Several visualization experiments are conducted to further study the effectiveness and limitations of enhancing DFER model with AU in detail.

\begin{figure*}
    \centering
    \includegraphics[width=.8\linewidth]{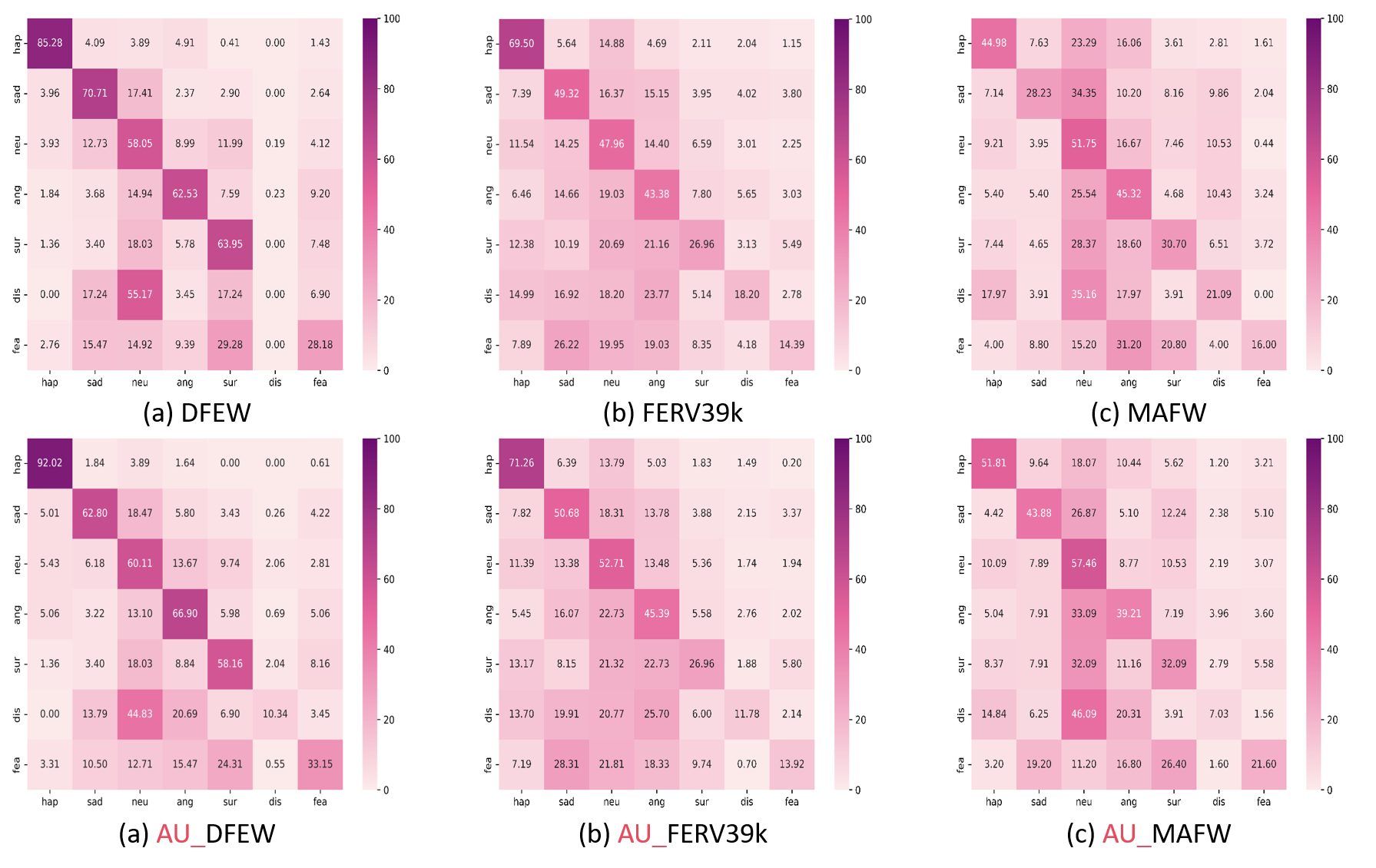}
  
    \caption{
   Heatmap of confusion matrix for M3DFEL in classification tasks, including: whether AU-enhanced, using on DFEW, FERV39k and MAFW datasets.}
    \label{fig:heatmap}
\end{figure*}

 \textbf{Heatmap.} We draw heatmaps to visualize the confusion matrix, and present them in Fig. \ref{fig:heatmap}. It can be seen from the figure that the improvement of AU-enhanced model is significant, especially for minor classes. Taking disgust samples in DFEW as example, the performance suffers from imbalanced data, while adding AU solves this problem to some extent. 

\begin{figure*}[!ht]
  \centering
  \includegraphics[width=.8\linewidth]{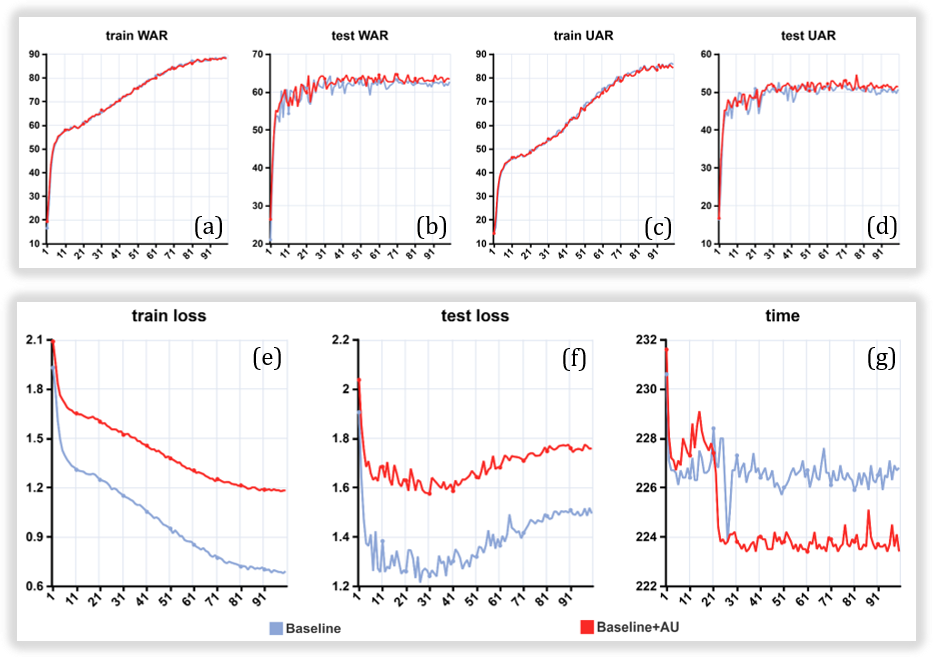}
  \caption{A comprehensive process for the utilization of AU expression knowledge in the training of M3DFEL on DFEW.}
  \label{fig:ProcessDatainDFEW}
\end{figure*}

\textbf{Visualization of training process.}
Applying AU-expression knowledge to M3DFEL with the knowledge injected as shown in Fig. \ref{fig:ProcessDatainDFEW}, test WAR(\ref{fig:ProcessDatainDFEW}b) and test UAR(\ref{fig:ProcessDatainDFEW}d) become higher. It is clear that after combining AU loss with expression loss, the total loss increased significantly(\ref{fig:ProcessDatainDFEW}f), because AU loss is high, and both decline at essentially the same rate. In addition, the time consumption for each epoch dropped dramatically after 30 epochs(\ref{fig:ProcessDatainDFEW}g), compared to the model without AU-expression knowledge, verifying that our method does not increase the computational cost.

\begin{figure}[!ht]
  \centering
  \includegraphics[width=.8\linewidth]{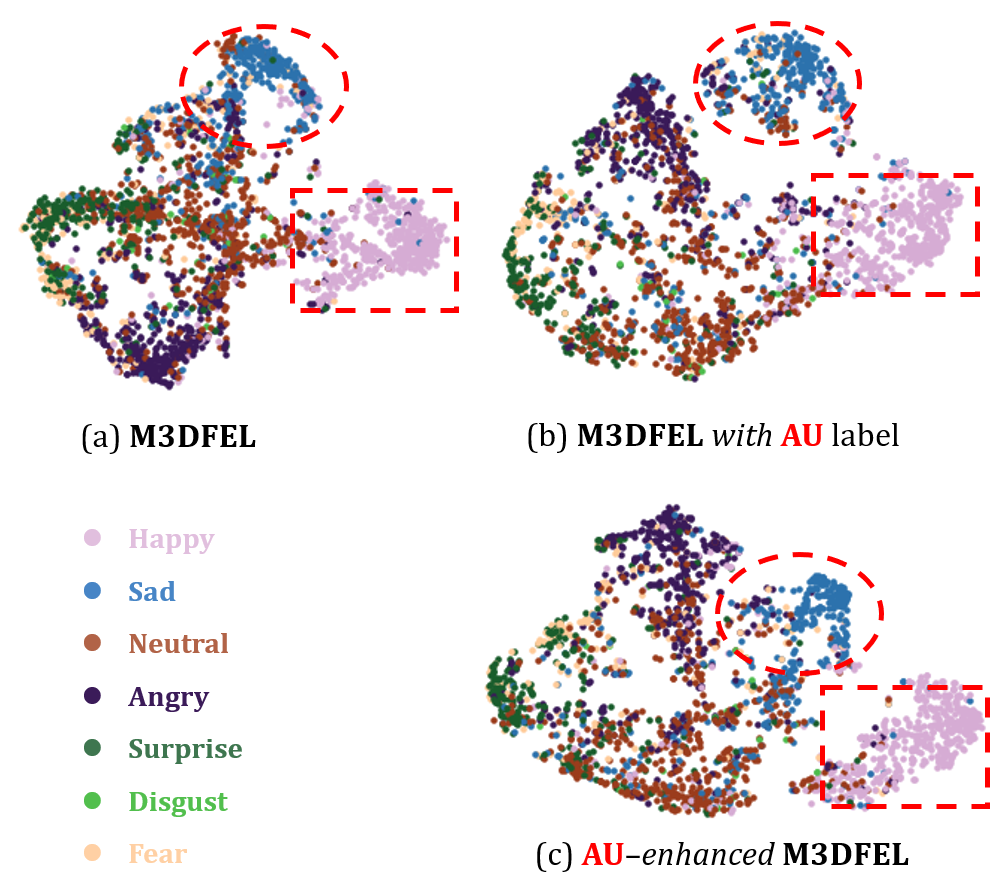}
  \caption{
  t-SNE visualization \cite{van2008visualizing} of dynamic facial expression features obtained by M3DFEL under different conditions: (a) only M3DFEL, (b) simply introduce AU in M3DFEL and (c) use AU in M3DFEL learning process. Different colors represent different expressions. The figure illustrates that AU can effectively improve the performance of DFER.}
  \label{fig:tsne}
  \vspace{-7pt}
\end{figure}

\textbf{t-SNE.} The t-SNE plot illustrates that adding AU-expression knowledge to the model can boost inter-class distance without increasing intra-class distance in Fig. \ref{fig:tsne}. What's more, neutral samples tend to overlap with all kinds of other samples, which suggests that neutral samples from macro-expression recognition can be considered as samples for micro-expression task.

\section{Discussion}
We conduct the AU-DFER rediscovery experiments using one DFER model, M3DFEL, which is replaceable, ensuring the flexibility of our architecture.

The results of the AU-DFER rediscovery experiments indicate that the AU distribution of neutral samples was more homogeneous than that of samples labeled with other expressions. Furthermore, inconsistency across datasets was observed, which suggests that neutral samples should be considered as samples with different micro-expressions in macro facial expression recognition. One of the key objectives of this research is to enhance the efficacy of DFER by integrating AUs in a systematic manner. 

However, the issue of data bias persists, and the correlation between AUs and expressions may be influenced by demographic characteristics and cultural backgrounds. Nevertheless, the current data and knowledge base is insufficient to fully address this challenge. Despite efforts to reduce annotation bias across the 4 datasets, the problem of data bias remains, and no generalized conclusions can be drawn to explain the correlation between AU and expressions. Even though OpenFace provides the most numbers detected AUs, the number is still limited, and the accuracy could be further improved, contributing to a limitation of our obtained AU-expression knowledge.

\section{Conclusion}
This study provides an in-depth analysis of facial muscle motor units and dynamic expressions, and proposes a learning framework paradigm, AU-DFER, based on AU prior knowledge injection to enhance dynamic facial expression recognition. The study acquired quantitative knowledge of AU expressions using 4 mainstream datasets and incorporated this into existing dynamic expression recognition methods in the form of AU loss, thereby achieving a significant improvement. The experiments corroborated the veracity of the AU quantitative knowledge and demonstrated that AU prior knowledge based on emotional psychology can enhance dynamic expression recognition. Furthermore, the study validates the effectiveness of diverse AU knowledge injection strategies across different models. Additionally, the experiments reinforce the comprehension of AU and highlight the potential of AU knowledge in the domain of dynamic expression recognition.

\section{Acknowledgment}
This work was supported by National Key Research and Development Program of China (2024YFC3606800) and by Shanghai Jiao Tong University 2030 Initiative.
\newpage
\bibliographystyle{ACM-Reference-Format}
\bibliography{sample-base}
\end{document}